\pdfoutput=1

\documentclass[11pt]{article}

\usepackage{coling}

\usepackage{times}
\usepackage{latexsym}

\usepackage[T1]{fontenc}

\usepackage[utf8]{inputenc}

\usepackage{microtype}

\usepackage{inconsolata}

\usepackage{graphicx}

\usepackage{array}
\usepackage{amsmath}
\usepackage{amsfonts}
\usepackage{hyperref}
\usepackage{placeins}
\usepackage{booktabs,arydshln}
\usepackage{multicol}
\usepackage[export]{adjustbox}
\usepackage{float} 

\makeatletter
\def\adl@drawiv#1#2#3{%
        \hskip.5\tabcolsep
        \xleaders#3{#2.5\@tempdimb #1{1}#2.5\@tempdimb}%
                #2\z@ plus1fil minus1fil\relax
        \hskip.5\tabcolsep}
\newcommand{\cdashlinelr}[1]{%
  \noalign{\vskip\aboverulesep
           \global\let\@dashdrawstore\adl@draw
           \global\let\adl@draw\adl@drawiv}
  \cdashline{#1}
  \noalign{\global\let\adl@draw\@dashdrawstore
           \vskip\belowrulesep}}
\makeatother

\title{Part-Of-Speech Sensitivity of Routers in Mixture of Experts Models}

\author{Elie Antoine\textsuperscript{1}\ , Frédéric Béchet\textsuperscript{1, 3}\ , Philippe Langlais\textsuperscript{2}\\
\textsuperscript{1}CNRS, LIS, Aix-Marseille Université, France \; \footnotesize{\texttt{\{first.last\}@lis-lab.fr}}\\
\textsuperscript{2}RALI, DIRO, Université de Montréal, Canada \; \footnotesize{\texttt{felipe@iro.umontreal.ca}}\\
\textsuperscript{3}International Laboratory on Learning Systems (ILLS - IRL CNRS), Montreal\\
}

\begin{document}
\maketitle
\begin{abstract}
This study investigates the behavior of model-integrated routers in Mixture of Experts (MoE) models, focusing on how tokens are routed based on their linguistic features, specifically Part-of-Speech (POS) tags. The goal is to explore across different MoE architectures whether experts specialize in processing tokens with similar linguistic traits. By analyzing token trajectories across experts and layers, we aim to uncover how MoE models handle linguistic information. Findings from six popular MoE models reveal expert specialization for specific POS categories, with routing paths showing high predictive accuracy for POS, highlighting the value of routing paths in characterizing tokens.
\end{abstract}

\section{Introduction}

Mixture of Experts (MoE) models, inspired by ensemble methods, offer an efficient parameter-to-performance ratio by partitioning the Feed Forward Network (FFN) layers into sub-groups of parameters called "\textit{experts}". A router model learns to direct each token to a subset of these experts, resulting in bare computation (or effective parameter count) that is significantly lower than what would be required for an equivalent dense model.

While many studies on MoE training include analyses of expert coactivations and potential specializations, often focusing on language or domain-specific tasks, comprehensive evaluations and comparisons of expert behavior across different MoE models remain scarce. These studies typically address specific aspects of MoE performance, leaving broader trends in expert behavior underexplored. Notably, related analyses of token routing, such as those in OpenMoE \cite{xue2024openmoe} and OLMoE \cite{muennighoff2024olmoe}, were part of broader investigations into model training. They revealed that tokens with the same token ID are often routed to the same expert regardless of context, suggesting inherent biases in routing mechanisms. Mixtral \cite{jiang2024mixtral} additionally hinted at syntactic specialization occurring, but this phenomenon has yet to be systematically studied across models. Similarly, \citet{zoph2022stmoedesigningstabletransferable} found that in an encoder-decoder network, specialization on syntactic features occurred only in the encoder experts, with no specialization in the decoder ones.

However, the specific types of specialization carried out by these experts, particularly from a linguistic perspective, and whether these specializations are consistent across different model architectures remain unclear.

We propose in this study to analyze routing decisions in open-weight MoE models to investigate whether experts specialize based on the syntactic nature of tokens represented by their parts-of-speech (POS) labels. By examining the top-$k$ experts chosen at each layer, we explore specialization in terms of tokens' POS labels both within individual layers and across the entire routing path of tokens. We use POS because they serve as the fundamental building block for the entire syntactic structure of a sentence. Our aim is thus to determine whether the routers acquired this crucial linguistic knowledge during the token processing.

We also leverage model-integrated routers from each layer as probing tools inspired by traditional probing methods such as the one of\;\citet{shi2016does} by examining the sequence of top-$k$ experts selected at each layer.
The goal of this study is to answer these two research questions:
\begin{itemize}
    \item \textbf{$Q_1$}: Are routers in Mixture of Experts models sensitive to the part of speech of tokens? 
    \item \textbf{$Q_2$}: Does this specialization occur in specific layers, or do the entire routing paths also encode syntactic information at the token level?
\end{itemize}

\section{Background on MoE architecture with Transformers}

A dense transformer \cite{vaswani2017attention} model is constructed by stacking $L$ layers of transformer blocks, each comprising a self-attention module and a FFN. MoE language models typically replace FFNs with MoE layers that consist of multiple "experts", which are smaller FFNs. In most cases, these experts are not trained to specialize in specific parts of the data; rather, they are subdivisions of the dense FFN layer into smaller networks aimed at computational efficiency. The top-$k$ routing mechanism, originally proposed by \citet{Shazeer2017OutrageouslyLN} is composed at each layer by a \textbf{router} (or gating network) which directs tokens to specific experts based on their relevance as follows. 

    \paragraph{The router is a learned linear layer} that maps input representations \( x \) to logits \( h(x) \) using a trainable weight matrix \( W_r \). This can be expressed as: 
    \vspace{-4mm}
    \[h(x) = W_r \cdot x\vspace{-2mm}\] 
    \paragraph{The logits \( h(x) \) are normalized} using a softmax function to produce routing probabilities \( p(x) \), ensuring they sum to one. The probability of routing token $x$ to the $i^{th}$ expert among $N$ is given by :
    \vspace{-1.5mm}
    \[p_i(x) = \frac{e^{h(x)_i}}{\sum_{j=1}^{N} e^{h(x)_j}}\vspace{-2mm}\] 
    \paragraph{The router selects the top-$k$ experts} based on the highest routing probabilities. This selection is implemented by taking the softmax over the top-$k$ logits of the linear layer. Formally, we define:
    \vspace{-1.5mm}
    \[G(x) := \text{Softmax}(\text{TopK}(x \cdot W_r))\vspace{-2mm}\]
    where $\text{TopK}(\ell)_i := \ell_i$  if $\ell_i$ is among the top-$k$ coordinates of logits $\ell \in \mathbb{R}^N$ and $-\infty$ otherwise.

    \paragraph{Each selected expert $E_i$} processes the input $x$. The output of each expert is then weighted by its respective routing probability $p_i(x)$.

    \paragraph{The final output of the MoE layer} is the weighted sum of the outputs from the selected experts where $T$ denotes the set of indices of the top-$k$ selected experts: $y = \sum_{i \in T} p_i(x) E_i(x)$

\vspace{0.2cm}

This mechanism allows for efficient and adaptive routing of tokens, using only the most relevant experts per token. The value of $k$ corresponding to the number of experts used per token, is a key hyperparameter that modulates the amount of computation required for processing each token. By increasing the total number of experts ($N$) while keeping $k$ fixed, one can expand the model’s parameter count (sparse parameter count) without significantly increasing computational cost, as only $k$ experts are active per token. Various optimizations, including sparse matrix multiplications and expert parallelism, can further enhance performance and manage GPU load balancing. For more details, see \cite{fedus2022reviewsparseexpertmodels}.

While classical MoE models do not have experts specifically trained for distinct data aspects, some research explored the concept of experts as fully independent models \cite{gururangan2021, li2022}. 
Leveraging modularity and composability allows experts to be trained on specific domains, thereby enabling the construction of custom networks from these specialized components. 

We are not interested in the latter, but want to analyse whether model experts who are not trained in a specific way for a sub-task show specialization. 

\begin{figure}[t]
    \centering
    \includegraphics[width=1\columnwidth]{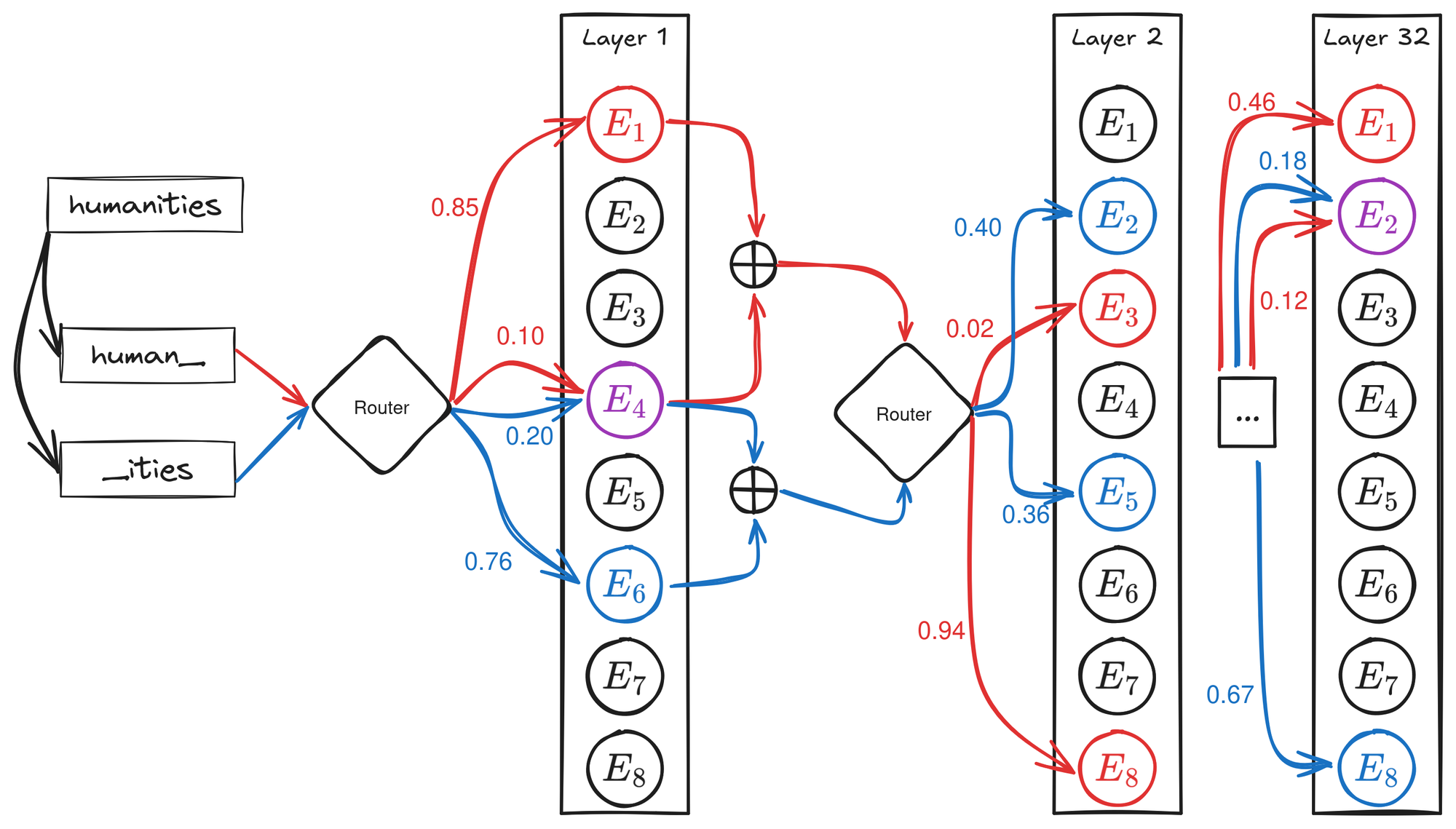}
    \caption{Example of token routing with 2 of 8 selected experts. For "human\_", the path is \([(1,4),(8,3),\ldots,(1,2)]\); for "\_ities", it is \([(6,4),(2,5),\ldots,(2,8)]\).}
    \label{fig:token_routing}
    \vspace{-2mm}
\end{figure}

\section{Methodology}

In this study, we utilize model-integrated routers primarily designed to direct tokens to relevant experts. Although their main function is to route tokens efficiently, we leverage these routers as probing tools to gain insights into the model behavior. These routers, already part of the trained model architecture, allow us to analyze the sequence of top-$k$ experts chosen at each layer.  By inputting sentences into various models, we record the sequence of experts assigned to each token per layer, as shown in Figure \ref{fig:token_routing}. The signals we use for our analysis are thus very light, in the worst case being of size $\mathcal{S}=\#route\; expert \times \#layer$.
For example, with \textbf{Mixtral-8x7B-v0.1} \cite{jiang2024mixtral}, a model of 32 layers having 2 of 8 experts routed for each token lead to a $32\times2$ signal matrix.

\subsection{Layer-wise Specialization Analysis}

We measure expert specialization at each layer \(l\) by first ordering experts based on how many tokens they handle for each specific POS. We then calculate the proportion of tokens handled by the top-$k$ experts of this ranking:
\[Spec_{\text{POS}, l} = \frac{T_{\text{POS},\text{top-}k}}{T_{\text{POS},all}} \times 100\]
where, for a given POS, \(T_{\text{POS},\text{top-}k}\) is the number of tokens processed by the top-$k$ experts that handle the most tokens and \(T_{\text{POS},\text{all}}\) represents the overall number of tokens for this POS. The top-$k$ experts are selected per layer, highlighting layer-specific routing dynamics.

The \textbf{average specialization score} for each POS across all layers is given by:
\[
Spec_{\text{POS}} = \frac{1}{L} \sum_{l=1}^L Spec_{\text{POS}, l}
\]
where $L$ is the number of layers.\\

The model's \textbf{global specialization score} is:

\[Spec = \frac{1}{13} \sum_{\text{POS}} Spec_{\text{POS}}\]

where \(13\) is the number of POS categories, excluding SYM and X which account for less than 1\% of the total, reflecting overall specialization consistency across all layers and POS categories\footnote{See Section \ref{sec:corpus} for the tagset being used.}.

We compare these values to the expected token routing percentages, denoted as $(\mathcal{U})$, which represent the proportion of tokens that top-$k$ experts would handle under a uniform distribution. The difference between $Spec$ and $\mathcal{U}$, denoted as $\Delta\;\mathcal{U}$, quantifies the deviation from this expectation.

\paragraph{POS Distribution as a Whole}\mbox{}

Layer-wise specialization of experts can also be assessed by comparing their POS distributions to that of the corpus. Greater divergence between these distributions indicates higher specialization, as experts deviate from simply mirroring the input distribution. To quantify this, we compute the Kullback-Leibler (KL) divergence between each expert's probability distribution at every layer and the corpus distribution.

We summarize this last score using three metrics: the average of either the mean (\textbf{$\mu$ Mean} ), maximum (\textbf{$\mu$ Max} ), or minimum (\textbf{$\mu$ Min}) KL divergence of all experts per layer, computed across all layers.

\subsection{Expert Routing Paths}
\label{sec:MLP_routing}
For each token we extract the $k$ routed experts at each layer to create a "path" of the token among the different experts of the model.

From this, we train a Multi-Layer Perceptron (MLP)\footnote{We used the \href{https://scikit-learn.org/stable/modules/generated/sklearn.neural_network.MLPClassifier.html}{MLPClassifier} of the Scikit-Learn library with the default parameters except for "max\_iter" which has been increased to ensure convergence.} to predict the POS of each token based on the expert routing information from each layer. 

The input to the MLP is the flattened vector \(\mathbf{r} \in \mathbb{R}^{S}\), ($S$ being 64 for Mixtral) which represents the list of experts chosen for the token at each layer. The target is a one-hot encoded POS vector \(\mathbf{y}\) with \(15\) possible classes. The MLP outputs a predicted vector \(\hat{\mathbf{y}}\) that represents the probabilities for each POS tag. We evaluate the routers' effectiveness by assessing the MLP's performance in predicting the correct POS on the test data, based on the expert routing paths.

\section{Experiment}

\subsection{Corpus}
\label{sec:corpus}
For the POS data, we use 5000 random English sentences from OntoNotes 5.0 \cite{weischedel2013ontonotes}, corresponding to 116,379 tokens according to the Mixtral tokenizer. This corpus has the advantage of including other types of annotations aligned with the POS, which can be used to expand the analysis we conduct here. We cast\footnote{We follow the conversion table given on the \href{https://universaldependencies.org/tagset-conversion/en-penn-uposf.html}{UD website}} the POS tags from Penn Treebank POS tags to Universal Dependencies (UD) tags, grouping them into broader categories such as all nouns, verbs, etc.

Tokens are linked to their word's POS by assigning each token the POS tag of the word it belongs to. For instance, in the case of "humanities," both tokens "human\_" and "\_ities" are tagged as NOUN.

The distribution of the proportion of POS in our subset is shown in Appendix~\ref{sec:appendix-POS}.

We split the data into a training set and a test set for the MLP training experiments described in Section~\ref{sec:MLP_routing} using a two-thirds/one-third split at the token level, allowing different tokens of the same word to appear in different sets.

\subsection{Models}

We compared 6 models in our experiments: \textbf{dbrx-base} (132B parameters, 36B active, 40 layers, 4 among 16 experts per layer \url{https://huggingface.co/databricks/dbrx-base}), \textbf{Mixtral-8x7B-v0.1} \citep[46.7B parameters, 13B active, 32 layers, 2 among 8 experts per layer][]{jiang2024mixtral}, \textbf{Phi-3.5-MoE-instruct} \citep[41.9B parameters, 6.6B active, 32 layers, 2 among 16 experts per layer][]{abdin2024phi3}, \textbf{deepseek-moe-16b-base} \citep[16.4B parameters, 2.8B active, 28 layers, 6 among 64 experts per layer, plus 2 shared experts][]{dai2024deepseekmoe}, \textbf{Qwen1.5-MoE-A2.7B} \citep[14.3B parameters, 2.7B active, 24 layers, 4 among 60 experts per layer][]{qwen_moe}, and \textbf{OLMoE-1B-7B} \citep[7B parameters, 1B active, 16 layers, 8 among 64 experts per layer][]{muennighoff2024olmoe}.

Among the 6 tested models, only OLMoE-1B-7B can be considered fully reproducible, as it is the only model with a documented and accessible training dataset alongside open-source code for both training and inference.

All models were loaded in fp8, and when a base version (i.e.\ without preference tuning) of the model existed, this was used. Experiments on Mixtral confirmed that this did not alter prediction results. Tests with full precision, half precision, and with/without the instruct version yielded nearly identical expert routing.

\section{Results}

\begin{table}[]
\centering
\setlength{\extrarowheight}{.5ex}
\resizebox{\columnwidth}{!}{%
\begin{tabular}{c|cc|c|c|c|}
\cline{2-6}
                                            & \multicolumn{2}{c|}{\textbf{MLP score}}    & \multicolumn{3}{c|}{\textbf{Specialization score}}  \\ \hline
\multicolumn{1}{|c|}{\textbf{model}}                & \multicolumn{1}{c|}{$top_k$} & $top_1$    & $Spec$    & $\mathcal{U}$ & $\Delta\;\mathcal{U}$\\ \hline
\multicolumn{1}{|c|}{dbrx-base}             & \multicolumn{1}{c|}{0.86}  & 0.83 & 51.87 & 25.0 & 26.87 \\ \hline
\multicolumn{1}{|c|}{Mixtral-8x7B-v0.1}     & \multicolumn{1}{c|}{0.84}  & 0.83 & 50.21 & 25.0 & 25.21 \\ \hline
\multicolumn{1}{|c|}{Phi-3.5-MoE-instruct}  & \multicolumn{1}{c|}{0.89}  & 0.88 & 48.49 & 12.5 & 35.99 \\ \hline
\multicolumn{1}{|c|}{deepseek-moe-16b-base} & \multicolumn{1}{c|}{0.80}  & 0.80 & 43.60 & 9.4  & 34.20 \\ \hline
\multicolumn{1}{|c|}{Qwen1.5-MoE-A2.7B}     & \multicolumn{1}{c|}{0.82}  & 0.79 & 38.85 & 6.7  & 32.15 \\ \hline
\multicolumn{1}{|c|}{OLMoE-1B-7B}           & \multicolumn{1}{c|}{0.75}  & 0.72 & 48.82 & 12.5 & 36.32 \\ \hline
\end{tabular}
}
\caption{MLP and global specialization scores ($Spec$) for each model. $top_k$ represents MLP accuracy with all experts, $top_1$ with the highest-probability expert. $(\mathcal{U})$ is the expected proportion of tokens under uniform distribution, and $\Delta\;\mathcal{U}$ is the deviation from it.}
\label{tab:all_score}
\end{table}

\subsection{Layer Wise Analysis}
If we look only at expert specialization, i.e. the percentage of tokens the top-$k$ experts retrieve from a given POS layer by layer, we observe a specialization. This approach ensures that we account for not only the top-1 logits but also consider the routing to multiple experts, reflecting the overall expert distribution. 
Table~\ref{tab:all_score} presents the overall scores for each model. All models exhibit a higher degree of specialization than expected with a uniform distribution, with $Spec_{POS}$ ranging from +25.21\% for Mixtral to +36.32\% for OlMoE, compared to the expected percentages shown in parentheses. 

These are average scores, with higher specialization observed at specific layers. Table \ref{tab:max_score_Phi}  highlights the maximum specialization $\max_{l} (Spec_{\text{POS},l})$ for 4 POS on the Phi-3.5-MoE-instruct model. Results for all POS and models are provided in Appendix~\ref{sec:appendix-detail-specialisation}.

\begin{table}[h]
\resizebox{\columnwidth}{!}{%
\begin{tabular}{|c|c|c|c|c|}
\hline
POS   & NOUN  & VERB  & PUNCT & ADJ   \\ \hline
$\max_{l} (Spec_{\text{POS},l})$ & 61.20 & 61.76 & 84.53 & 58.39 \\ \hline
$\Delta U$ & +48.70 & +49.26 & +72.03 & +45.89 \\ \hline
\end{tabular}
}
\caption{$\max_{l} Spec_{\text{POS},l}$ across layers for Phi-3.5-MoE-instruct on specific POS, $\Delta U$ showing deviation from the uniform distribution.}
\label{tab:max_score_Phi}
\end{table}

Table~\ref{tab:kl_statistics} summarizes the KL divergence statistics across models. dbrx-base shows the lowest $\mu$ Mean (0.21), while Qwen shows the highest $\mu$ Max (1.92), suggesting stronger expert specialization in certain layers.  Notably, OLMoE maintains a fairly balanced profile with low $\mu$ Min (0.04) but higher $\mu$ Max (1.52), reflecting both shared and distinct expert behaviors (see Appendix \ref{sec:appendix-KL_heatmap} for detailed KL at each expert and layer). However, KL divergence is less interpretable than the other metrics due to its unbounded scale $(0,\infty)$ and the difficulty in defining what represents a small or large divergence.

\begin{table}[]
\resizebox{\columnwidth}{!}{%
\centering
\begin{tabular}{|c|c|c|c|}
\hline
\textbf{model} & \textbf{$\mu$ Min} & \textbf{$\mu$ Max} & \textbf{$\mu$ Mean} \\
\hline
dbrx-base & 0.047 & 0.55 & 0.21 \\
\hline
Mixtral-8x7B-v0.1 & 0.11 & 0.40 & 0.23 \\
\hline
Phi-3.5-MoE-instruct & 0.14 & 1.49 & 0.60 \\
\hline
deepseek-moe-16b-base & 0.10 & 1.73 & 0.60 \\
\hline
Qwen1.5-MoE-A2.7B & 0.16 & 1.92 & 0.73 \\
\hline
OLMoE-1B-7B & 0.04 & 1.52 & 0.53 \\
\hline
\end{tabular}
}
\caption{Kullback-Leibler divergence statistics}
\label{tab:kl_statistics}
\end{table}

\subsection{POS Prediction Using Expert Paths}

\begin{figure*}[ht]
  \includegraphics[width=1\textwidth,center]{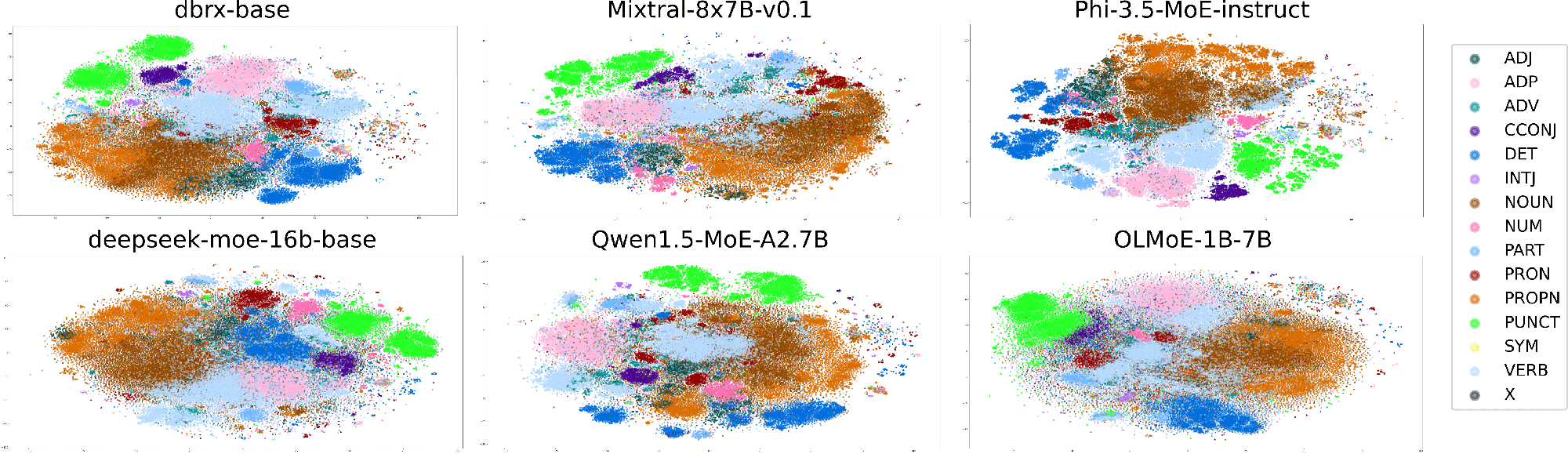}
  \caption{2D-TSNE projection of token path}
  \label{fig:TSNE}
\end{figure*}

The results from MLPs trained with paths from different models are summarized in Table~\ref{tab:all_score}. All models show an accuracy between 0.79 and 0.88, except OlMoE, regardless of whether full routing information or just the top-expert information from the gating network is used. A simple baseline, predicting the most common POS for a word form, achieves an accuracy of 0.91. However, predicting POS from token-level routing paths is more challenging, as it relies solely on how the token was routed, without any explicit information about the token's form. Despite this, MLPs still perform well, suggesting that the router effectively captures syntactic information.

Examining the confusion matrices (see Appendix~\ref{sec:appendix-confusion-matrices} for all matrices) on the test set reveals which POS types are better predicted, indicating stronger clustering in the paths and experts used. This analysis highlights connections between categories; for example, symbols (SYM) are often confused with punctuation (PUNCT) or numbers (NUM), and adverbs (ADV) and adjectives (ADJ) with nouns (NOUN) or verbs (VERB). Notably, the matrix for OLMoE is more distinct, showing significantly greater confusion across classes such as SYM, PUNCT, and ADJ and an overall lower average accuracy compared to the other models.

\paragraph{\textbf{Path Clustering Visualization Using TSNE}}\mbox{}\\
The TSNE visualization of token paths across different models (see Figure~\ref{fig:TSNE}) highlights clear and distinct clustering patterns for most POS categories, such as nouns, verbs, adjectives, and punctuation, for all models. These clusters demonstrate the models' ability to route tokens into syntactically coherent groups with minimal overlap. However, the visualization for \textit{OLMoE-1B-7B} reveals more diffuse and overlapping clusters, particularly among symbols, punctuation, and adjectives. This aligns with the confusion matrix analysis, which indicated greater class confusion and lower overall accuracy.

\paragraph{\textbf{Ablation Study}}\mbox{}\\
We conducted an ablation study to identify which layers encode the most POS-related information by training classifiers while progressively removing information from either the first or last layers of the model.  As shown in Figure~\ref{fig:ablation_MLP_Mixtral}, removing information from the first layers has a greater impact on POS prediction compared to removing information from the last layers for dbrx-base, deepseek-moe and Phi-3.5-MoE-instruct. This suggests that for these models, the earlier layers contain more token-characterizing information.

\begin{figure}[h!]
  \includegraphics[width=1\columnwidth]{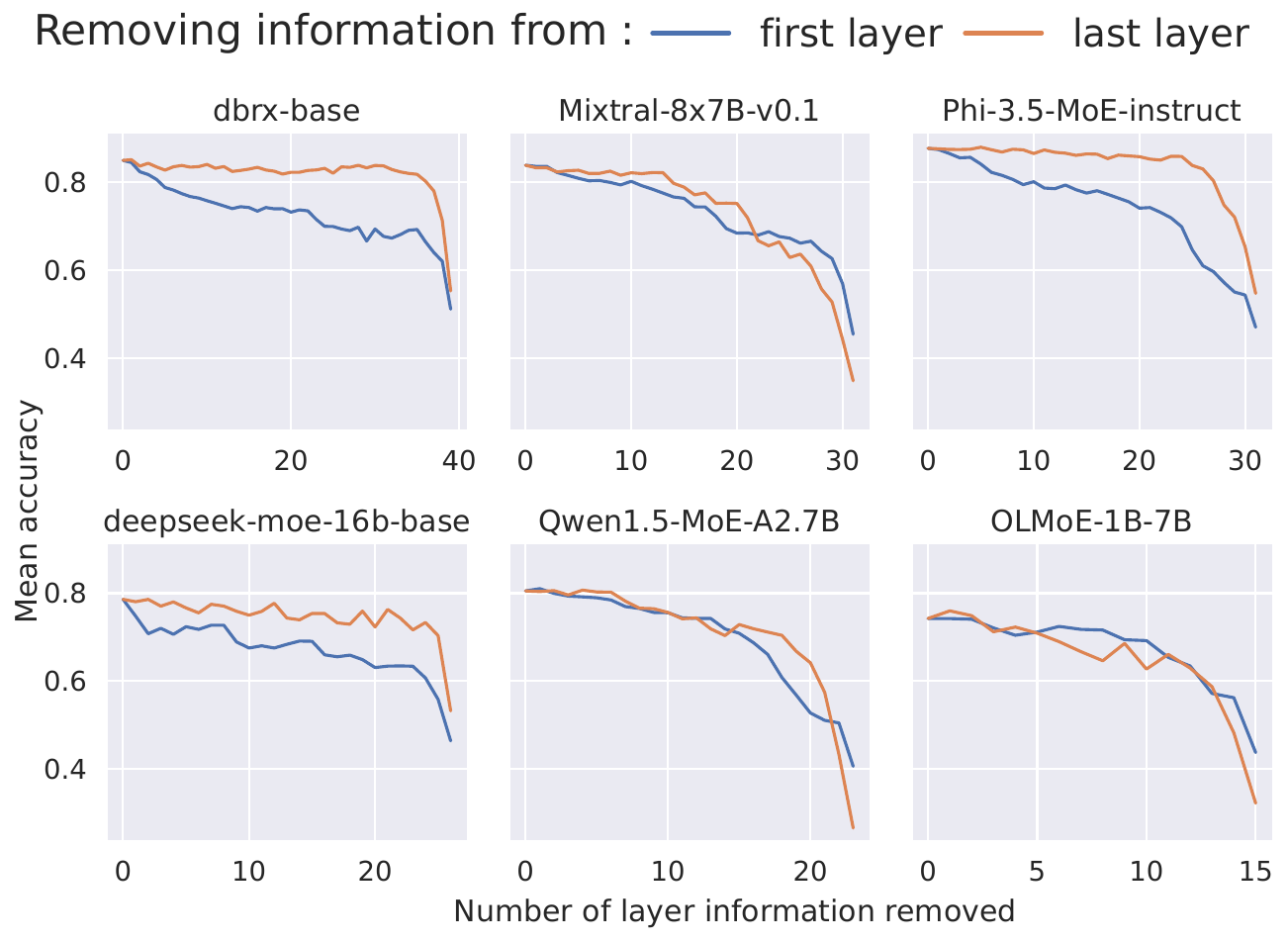}
  \caption {Accuracy of MLP trained on ablated signal per model, removing information from first or last layers.}
  \label{fig:ablation_MLP_Mixtral}
\end{figure}

\section{Conclusion}

This study explores the behavior of model-integrated routers in various MoE models, focusing on token routing based on POS. By tracking the sequence of experts assigned to each token at each layer, we analyzed how these routers impact the model's processing strategy.
A key finding is the \textbf{specialization of experts for specific POS categories} ($Q_1$). Certain tokens are consistently routed to a few experts, this specialization being more pronounced for symbols and punctuation tokens.
Additionally, using MLPs to predict POS from routing paths showed high predictive accuracy across most models, indicating that \textbf{routing paths contain significant information about token characteristics} for many current MoE architectures ($Q_2$).

\newpage
\section{Limitation}
A limitation of our study is the focus on context tokens, without examining router behavior on generated tokens. Additionally, we only tested English within the domain covered by OntoNotes, which consists of relatively short sentences.

\section*{Acknowledgements}
This project was provided with computer and storage resources by GENCI at IDRIS thanks to the grant 2023-AD011012688R2 on the supercomputer Jean Zay's V100/A100 partition. We would like to thank the reviewers for their useful comments and feedback.

\bibliography{custom}

\clearpage
\onecolumn
\appendix
\FloatBarrier
\section{Proportion of POS}
\label{sec:appendix-POS}
\begin{table}[H]
    \centering
    \resizebox{0.25\columnwidth}{!}{
    \begin{tabular}{|c|c|c|c|}
        \hline
        POS & Count & \% of Total \\
        \hline
        SYM & 82 & 0.07\% \\
        X & 84 & 0.07\% \\
        INTJ & 1347 & 1.16\% \\
        PART & 2675 & 2.30\% \\
        CCONJ & 2760 & 2.37\% \\
        NUM & 2791 & 2.40\% \\
        PRON & 4435 & 3.81\% \\
        ADV & 5530 & 4.75\% \\
        ADJ & 7125 & 6.12\% \\
        PUNCT & 11237 & 9.66\% \\
        DET & 11695 & 10.05\% \\
        ADP & 11982 & 10.30\% \\
        PROPN & 15547 & 13.36\% \\
        VERB & 18091 & 15.54\% \\
        NOUN & 20998 & 18.04\% \\
        \hline
    \end{tabular}}
    \caption{POS Counts and Percentages}
    \label{tab:poSpec_counts}
\end{table}
\FloatBarrier

\section{MLP's confusion matrix for all models}
\label{sec:appendix-confusion-matrices}
\FloatBarrier
\begin{figure}[H]
  \includegraphics[width=1\textwidth]{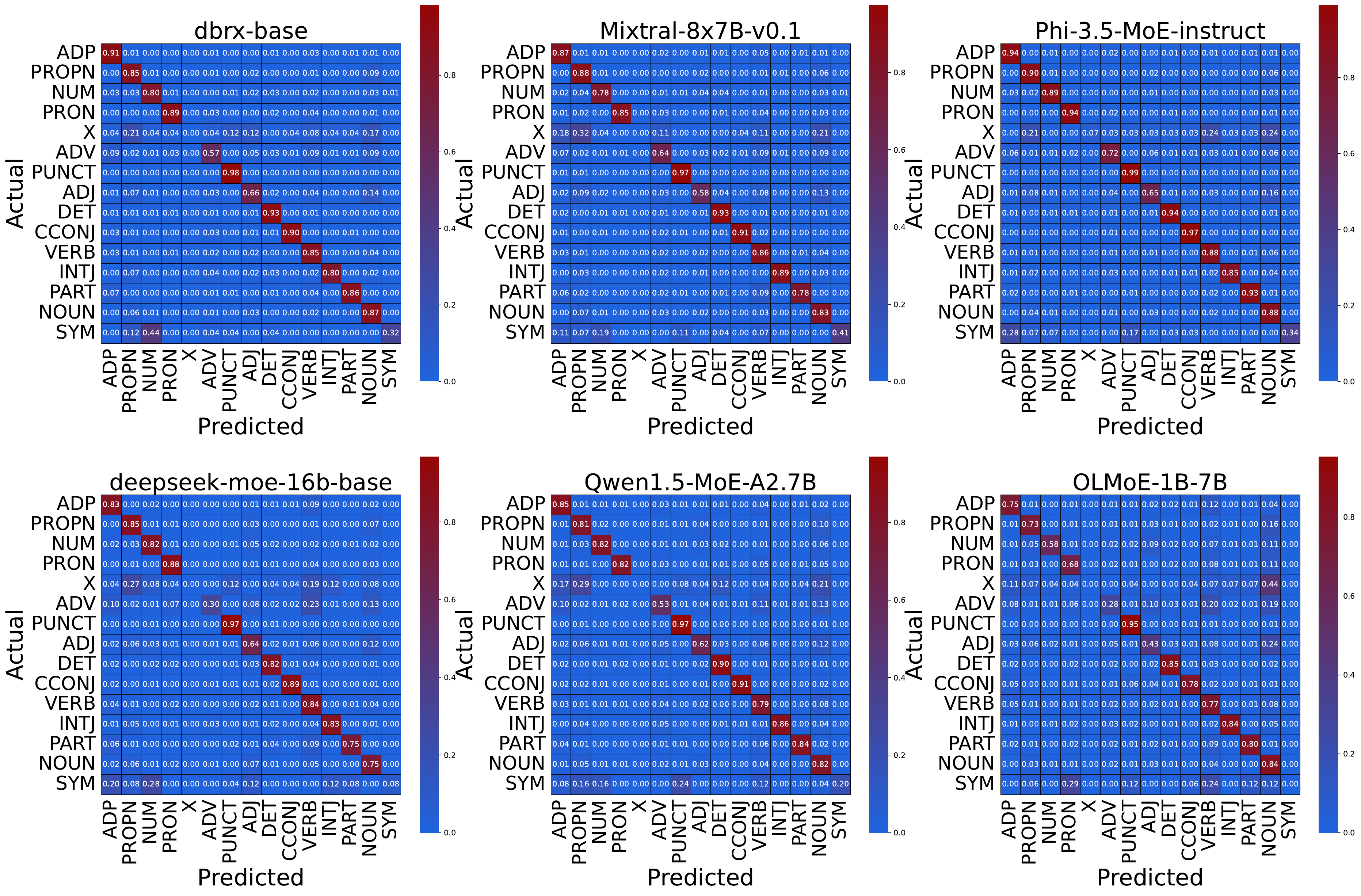}
  \caption {MLP's confusion matrix on the POS for all models}
  \label{fig:ablation_MLP_Mixtral}
\end{figure}

\FloatBarrier
\section{Detailed $Spec_{POS}$ for all models}
\FloatBarrier
\label{sec:appendix-detail-specialisation}
\FloatBarrier
\begin{table}[H]
    \centering
    \resizebox{\columnwidth}{!}{
    \begin{tabular}{lcccccc}
        \toprule
        Token   & dbrx-base @4/16 & Mixtral @2/8 & Phi-3.5-MoE-instruct @2/16 & deepseek-moebase @6/64 & Qwen-A2 @4/60 & OLMoE-1B-7B-0924 @8/64 \\
        \midrule
        ADP     & 52.19 (68.27)       & 52.45 (77.43)      & 51.69 (72.33)      & 44.99 (54.67)      & 39.39 (52.08)    & 50.80 (64.74)    \\
        PROPN   & 50.58 (59.17)       & 42.38 (66.29)      & 49.62 (64.89)      & 43.57 (53.02)      & 36.00 (51.83)    & 47.91 (56.61)    \\
        NUM     & 49.83 (58.91)       & 48.92 (69.94)      & 50.17 (68.28)      & 48.79 (57.90)      & 43.54 (54.52)    & 51.22 (61.79)    \\
        PRON    & 50.46 (66.14)       & 52.55 (80.30)      & 48.27 (73.54)      & 43.28 (57.39)      & 37.50 (52.29)    & 44.16 (57.31)    \\
        X       & 44.73 (57.88)       & 39.88 (65.48)      & 32.05 (45.51)      & 31.67 (37.39)      & 24.80 (34.53)    & 34.94 (41.49)    \\
        ADV     & 42.85 (53.27)       & 41.41 (57.06)      & 40.54 (60.50)      & 31.54 (43.25)      & 27.28 (38.05)    & 36.27 (41.26)    \\
        PUNCT   & 66.33 (80.56)       & 63.26 (81.97)      & 64.18 (84.53)      & 62.15 (75.49)      & 57.13 (73.90)    & 65.21 (72.81)    \\
        ADJ     & 47.57 (60.03)       & 42.59 (59.42)      & 39.38 (58.39)      & 36.06 (45.72)      & 32.50 (43.80)    & 43.26 (57.13)    \\
        DET     & 51.73 (70.22)       & 54.77 (87.37)      & 52.46 (71.89)      & 43.15 (67.82)      & 43.64 (58.18)    & 54.65 (66.78)    \\
        CCONJ   & 56.73 (77.58)       & 55.30 (78.42)      & 55.48 (87.43)      & 49.45 (63.89)      & 49.05 (67.37)    & 56.12 (68.87)    \\
        VERB    & 45.10 (56.23)       & 42.96 (64.67)      & 42.74 (61.76)      & 33.73 (46.25)      & 29.27 (43.67)    & 39.44 (43.56)    \\
        INTJ    & 59.39 (77.00)       & 56.70 (76.91)      & 48.38 (63.53)      & 54.29 (62.22)      & 45.31 (61.31)    & 53.08 (62.41)    \\
        PART    & 50.40 (58.88)       & 51.63 (69.53)      & 42.32 (66.47)      & 39.22 (49.59)      & 36.76 (48.84)    & 45.23 (53.00)    \\
        NOUN    & 51.12 (68.27)       & 47.83 (70.75)      & 45.11 (61.20)      & 36.67 (43.60)      & 27.66 (32.35)    & 47.37 (56.17)    \\
        SYM     & 62.50 (78.90)       & 58.25 (76.83)      & 40.36 (85.63)      & 42.97 (72.29)      & 30.07 (33.83)    & 45.78 (54.56)    \\
        \cdashlinelr{1-7}
        $(\mathcal{U})$ & 25.0        & 25.0               & 12.5               & 9.4                & 6.7              & 12.5             \\
        Mean ($Spec$) & 51.87      & 50.21              & 48.49              & 43.60              & 38.85            & 48.82            \\
        \bottomrule
    \end{tabular}
}
    \caption{Mean specialization ($Spec_{POS}$) of all layers per POS category: percentage of tokens routed by top k (@k) experts for each model. In parenthesis is the $\max_{l} Spec_{\text{POS},l}$ for the POS. $(\mathcal{U})$ shows the expected percentage of tokens recovered by top-$k$ experts under uniform token distribution. Model’s global specialization ($Spec$) is computed without taking X and SYM into account.}
    \label{tab:detail_expert_Mixtral}
\end{table}
\FloatBarrier

\FloatBarrier

\section{KL divergence matrices for all models}
\label{sec:appendix-KL_heatmap}
\FloatBarrier
\begin{figure}[H]
  \includegraphics[width=1\textwidth]{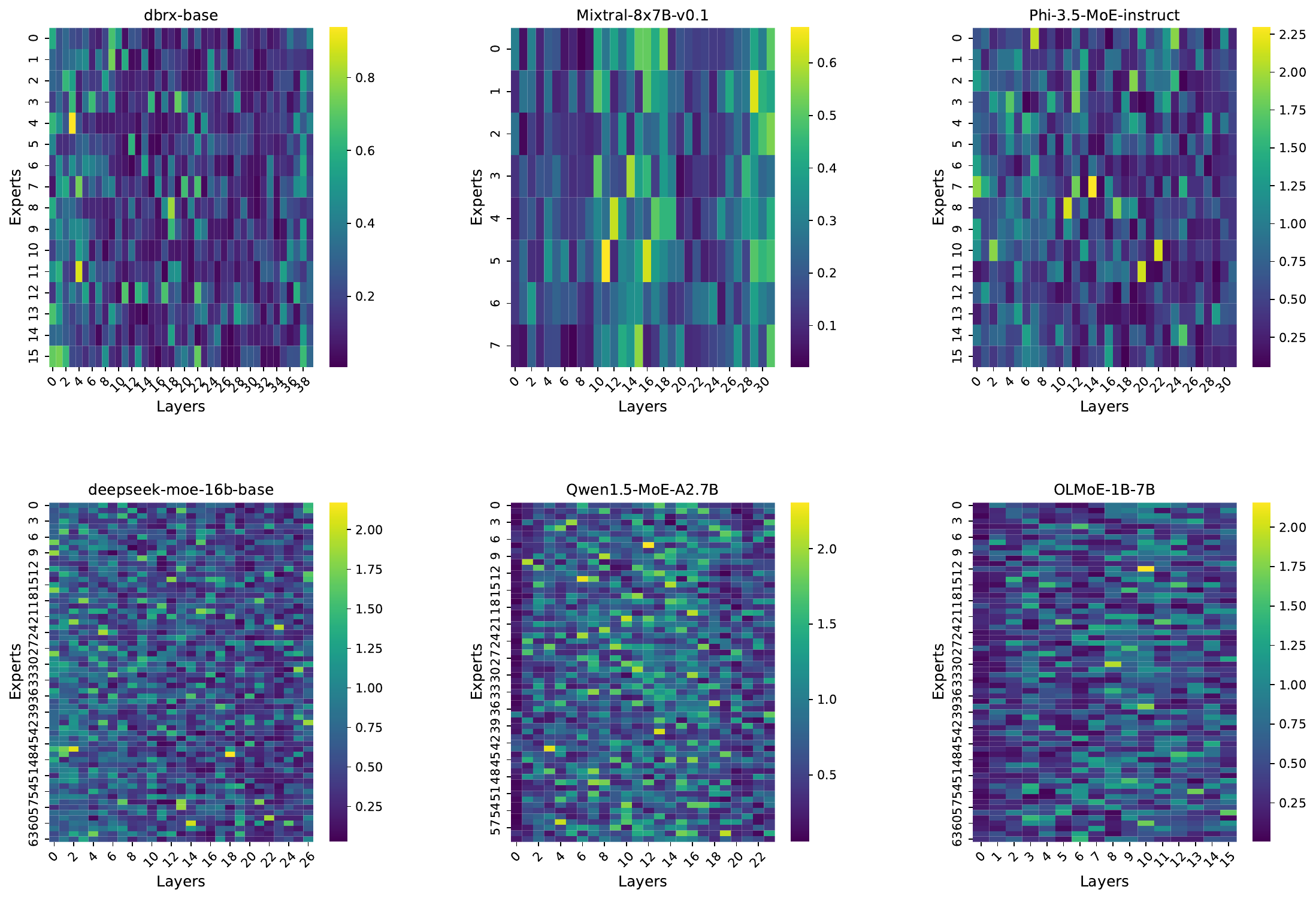}
  \caption {KL Divergence Matrices for all Models. Heatmaps showing the KL divergence between expert distributions and uniform distribution at each layer. The x-axis represents the layers, the y-axis represents the experts, and the color scale indicates the magnitude of KL divergence.}
  \label{fig:KL_heatmap}
\end{figure}
\FloatBarrier

\end{document}